# Spectroscopic Photoacoustic Denoising Framework using Hybrid Analytical and Data-free Learning Method

Fangzhou Lin[a,†], Shang Gao[a,†], Yichuan Tang[a], Xihan Ma[a], Ryo Murakami[a], Ziming Zhang[b], John D. Obayemi[c,d], Winston O. Soboyejo[d], Haichong K. Zhang[a,c,e]*

**Affiliations**
[a] Department of Robotics Engineering, Worcester Polytechnic Institute, 100 Institute Road, Worcester, MA 01609, USA
[b] Department of Electrical & Computer Engineering, Worcester Polytechnic Institute, 100 Institute Road, Worcester, MA 01609, USA
[c] Department of Biomedical Engineering, Gateway Park Life Sciences Center, Worcester Polytechnic Institute (WPI), 60 Prescott Street, Worcester, MA, 01605, USA
[d] Department of Mechanical & Materials Engineering, Worcester Polytechnic Institute, 100 Institute Road, Worcester, MA 01609, USA
[e] Department of Computer Science, Worcester Polytechnic Institute, 100 Institute Road, Worcester, MA 01609, USA

[*] Corresponding author: Haichong K. Zhang, hzhang10@wpi.edu
[†] The two authors contributed equally to this work.

**Abstract:** Spectroscopic photoacoustic (sPA) imaging uses multiple wavelengths to differentiate and quantify chromophores based on their unique optical absorption spectra. This technique has been widely applied in areas such as vascular mapping, tumor detection, and therapeutic monitoring. However, PA imaging is highly susceptible to noise, leading to a low signal-to-noise ratio (SNR) and compromised image quality. Furthermore, low SNR in spectral data adversely affects spectral unmixing outcomes, hindering accurate quantitative PA imaging. Traditional denoising techniques like frame averaging, though effective in improving SNR, can be impractical for dynamic imaging scenarios due to reduced frame rates. Advanced methods, including learning-based approaches and analytical algorithms, have demonstrated promise but often require extensive training data and parameter tuning. Moreover, spectral information preservation is unclear, limiting their adaptability for clinical usage. Additionally, training data is not always accessible for learning-based methods. In this work, we propose a Spectroscopic Photoacoustic Denoising (SPADE) framework using hybrid analytical and data-free learning method. This framework integrates a data-free learning-based method with an efficient BM3D-based analytical approach while preserving spectral integrity, providing noise reduction, and ensuring that functional information is maintained. The SPADE framework was validated through simulation, phantom, in vivo, and ex vivo studies. These studies demonstrated that SPADE improved image SNR by over 15 dB in high noise cases and preserved spectral information (R > 0.8), outperforming conventional methods, especially in low SNR conditions. SPADE presents a promising solution for preserving the accuracy of quantitative PA imaging in clinical applications where noise reduction and spectral preservation are critical.

## 1. Introduction

Photoacoustic (PA) imaging is an emerging functional imaging modality that relies on laser-generated ultrasound (US). Chromophores absorb a short light pulse and generate thermoelastic waves through adiabatic expansion, revealing the optical absorption of the material [1,2]. Different materials possess unique spectroscopic characteristics (i.e. PA spectrum). Spectroscopic PA (sPA) imaging, also known as multispectral PA imaging, can be used to characterize different types of chromophores using multiwavelength PA excitation [3]. Quantitative PA imaging leverages this spectral information to estimate the concentration of specific chromophores, allowing for more precise assessments in applications such as neurovascular mapping [4,5], contrast agent quantification for tumor detection [6–8], oxygenation mapping [9,10], and therapeutic monitoring [11–15], thereby enhancing diagnostic accuracy and treatment evaluation.

While it has demonstrated significant promise in many applications, PA imaging is susceptible to noise and artifacts. Noises arise from various sources including acoustic background noise, as well as electronic noise from detectors and amplifiers [16–19]}. These different noises in the PA signal led to a low signal-to-noise ratio (SNR), resulting in poor PA imaging quality. In order to manage noise signals and provide higher contrast PA imaging, denoising is performed to enhance the quality of this imaging modality. Averaging is a simple and effective technique for enhancing SNR by suppressing random noise and perservering spectrum information. It usually results in a lower frame rate and is not always feasible in dynamic scenarios with motion.

Several advanced denoising methods that require fewer imaging frames have been reported. Most of these methods can be categorized as data-driven learning-based methods and analytical methods. On the one hand, data-driven learning-based methods achieve remarkable progress in denoising various medical imaging [18,20]. He *et al.* designed an Attentive GAN for denoising PA microscopy imaging [21]. Although this method is promising for different datasets, it highly relies on the collection and annotation of datasets, which are not widely available in PA imaging [22]. Cheng *et al.* extend learning-based denoising to an unsupervised method based on the Noise2Noise network, which learns noise patterns from noisy image datasets [20]. However, ground truth images are not always available for supervised learning, and training data may be inaccessible for unsupervised learning, particularly in specific quantitative PA imaging tasks for clinical applications. Additionally, networks in data-driven learning may require additional training for new tasks to avoid performance drops due to the data distribution gap [23–25]. This limits the learning-based methods to work only in some specific tasks and hinders their genericity. On the other hand, numerous analytical algorithms have been introduced for denoising PA signals in single and multiple image frames [26]. Awasthi *et al.* proposed image-guided filtering for improving PA tomographic image reconstruction [27]. This guided filtering approach was proposed by combining the best features in the input and support images. Although it can improve the reconstructed image quality substantially, the supported images for feature enhancement cannot always be obtained. Kong *et al.* proposed an empirical mode decomposition for image reconstruction and quality improvement of PA imaging [28]. Despite the fact that both artificial and stochastic noises can be reduced, it needs parameter tuning with an optimized solution. Shi *et al.* designed a spatiotemporal singular value decomposition (SVD) for low fluency LED-based PA denoising [29]. Although this method shows great promise in enhancing imaging quality, they only demonstrated the effectiveness on single wavelength PA images without demonstration on sPA images. Kazakeviciute *et al.* provided effective solutions for artifact removal and denoising in multispectral PA images [30]. Their approach significantly improved the image clarity, yet focused more on spatial noise reduction without prioritizing spectral information preservation. The spectral integrity of the output multiwavelength PA image is unclear, and this hinders their applications in the sPA imaging denoise task. Having an advanced annotation-free and data-free (zero-shot fashion) denoising algorithm that could preserve spectral information is critical for sPA imaging.

In this work, we proposed a sPA denoising framework using a hybrid analytical and data-free learning method. A data-free learning-based method has recently been proposed which only consumes input data for imaging denoising [31–33]. The proposed method does not require additional images for network training or extensive hyperparameter tuning. For data preprocessing, we provide an explicitly Spectral Domain Data Re-assembly (SDDR) module, which is especially for consuming sPA imaging with multispectral. In the learning-based component, we adapt residual learning with redesigned Zero-Shot Noise2Noise (ZS-N2N). This model focuses on predicting noise rather than mapping the entire image. Doing so simplifies the learning task and helps preserve functional information. In the analytical component, we design a simple but effective BM3D-based method, which can provide stable denoising performance in sPA imaging with low computational cost. Together, we form a framework that effectively and efficiently denoises sPA images while preserving their spectrum information.

We summarize our contributions as follows: (1) We propose a SPADE framework, to denoise sPA images without training data and parameter tuning. Working in a zero-shot, unsupervised fashion, our framework offers broad applicability and adaptability to various quantitative PA imaging scenarios. (2) The framework preserves the spectral integrity of the pixel during the denoising allowing its usage for functional imaging. (3) We demonstrate notable denoising performance on simulation, phantom, ex vivo, and in vivo setup. The remainder of the paper was structured as follows. Section 2 details the proposed denoising method and algorithm. Section 3 outlines the experiment setup to validate the proposed denoising method in simulation, phantom, ex vivo, and in vivo setup. Section 4 presents the results of the validation experiments. The advantages and limitations of the work are discussed in Section 5, followed by the conclusion in Section 6.

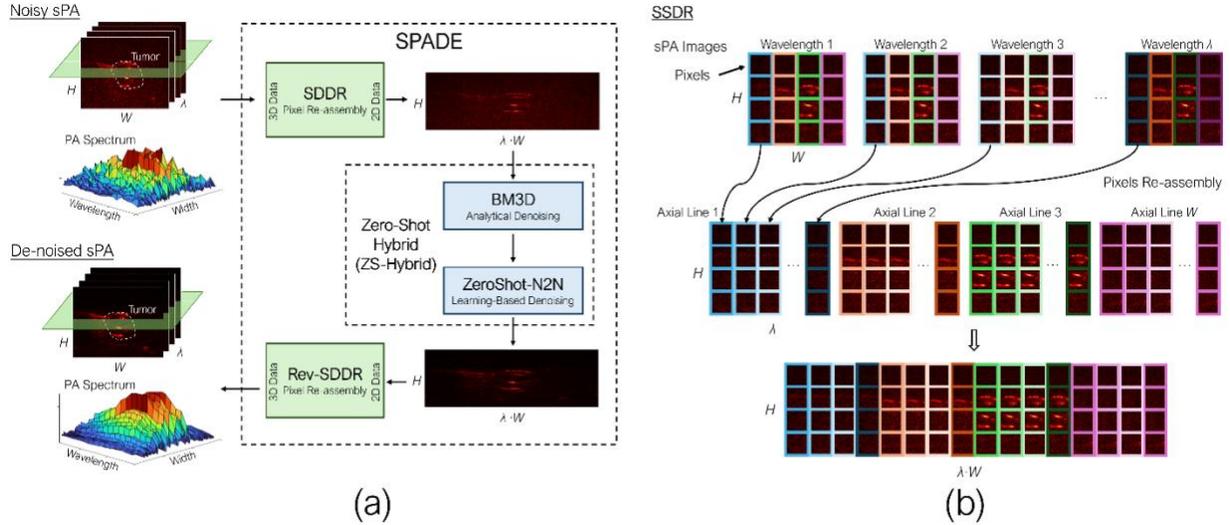

Figure 1. (a) Spectroscopic Photoacoustic Denoising (SPADE) Framework. Data processing pipeline of the proposed SPADE method: The noisy spectroscopic photoacoustic (sPA) image is first reassembled using Spectral Domain Data Re-assembly (SDDR) before being denoised as a 2D image using Zero-Shot Hybrid (ZS-Hybrid). After denoising, the reverse SDDR is applied to restore the original dimensions of the sPA image. (b) Illustration of sPA image pixel reassembly during the SDDR process. Box color denotes the pixel wavelength, and brightness indicates the axial location.

2. Methods

The proposed SPADE framework consists of two components: an SSDR module to directly multispectral sPA imaging for denoising, and Zero-Shot Hybrid (ZS-Hybrid) module leverage redesigned BM3D [34] and ZS-N2N [31] for two-stage sPA imaging denoising. The framework and image processing pipeline are shown in Fig. 1(a). Notably, the proposed framework consumes noise-included sPA only and the ZS-Hybrid network is trained in a zero-shot learning fashion.

    a. **Spectral Domain Data Re-assembly (SDDR) for Multispectral sPA Imaging**

In multispectral PA imaging, each pixel in a PA image collected at a specific wavelength has a unique spectral intensity while maintaining similar spatial information. This allows the use of multiple images for the same region. To format these regions, we designed a novel SDDR module as the preprocessing backbone for our SPADE. The noisy sPA image data is first processed in the SDDR module before being fed into the denoising model. The design groups pixels from the same position to ensure they share spatial information but differ in spectral information, which helps generate pairs of downsampled images in the following processing steps and enhances residual learning. The goal is to group pixels with different spectra horizontally, and then stack these groups hierarchically, preserving spectral information without compromising spatial integrity.

The SDDR pipeline is illustrated in Fig. 1(b). In each sPA image, there are $\lambda$ spectral frames with dimensions $W \times H$. The SDDR module consists of three steps: First, each axial line of pixels is split within each wavelength. Next, these axial lines are recursively concatenated horizontally to form a new 2D image with a size of $H\lambda$. This process is repeated for every axial line across the lateral axis, resulting in grouped axial-spectral plane pseudo-images. Finally, all these images are stacked laterally to create a new 2D image y with dimensions $H \times \lambda W$.

### b. Zero Shot-Denoising (ZED) for sPA denoising

Our proposed denoising method combines both analytical and data-free learning-based approaches. In the analytical component, we adapted the concept of the BM3D algorithm. Vanilla BM3D has become a benchmark in the medical imaging field due to its reasonable computational cost and consistent performance. It works by constructing a 3D array by aligning similar blocks extracted from noisy blocks within an image and then reintegrating these processed patches to reduce noise. The method leverages the inherent redundancies in images, following a two-step process: first, it applies hard thresholding during collaborative filtering to obtain a basic estimate (hard parameters), then applies a Wiener filter on the original noisy image and the basic estimate to generate the final denoised image [34]. In our method, we extend vanilla BM3D to spectral-BM3D, which accommodates the multiwavelength PA images processed by the SDDR module. Using vanilla BM3D with the SDDR module directly would result in high computational costs and disrupt spectral information, while spectral-BM3D aligns similar blocks in each grouped image, reducing processing time and preserving spectral information.

For the data-free learning-based component, we adapted ZS-N2N, a zero-shot image denoising algorithm that requires no training examples or prior knowledge of the noise model or level. ZS-N2N builds on Noise2Noise (NB2NB) [32] and Neighbour2Neighbour [33], where Noise2Noise trains on pairs of noisy images, and NB2NB generates such pairs from a single noisy image. ZS-N2N denoises by downsampling a single image into two pairs of images, then training a neural network to map one downsampled image to the other [31]. The downsampled images are generated by applying two convolution masks to the noisy image, as shown in Eq. 1.

$$D_1(y) = yk_1, \quad D_2(y) = yk_2, \quad (1)$$

where $k_1 = \begin{bmatrix} 0 & 0.5 \\ 0.5 & 0 \end{bmatrix}$, $k_2 = \begin{bmatrix} 0.5 & 0 \\ 0 & 0.5 \end{bmatrix}$. $D_1$ and $D_2$ are downsampled original image y, using fixed filter $k_1$, and $k_2$, respectively. The neural network is trained by minimizing two types of losses, residual loss ($\mathcal{L}_{res.}(\theta)$) and consistency loss ($\mathcal{L}_{con.}(\theta)$), using stochastic gradient descent as shown in Eq. 2. θ represents the parameters of the network.

$$\mathcal{L}(\theta) = \mathcal{L}_{res.}(\theta) + \mathcal{L}_{con.}(\theta) \quad (2)$$

The residual loss and consistency loss are calculated using Eq. 3 and Eq. 4, where $f_\theta$ is the network. ZS-N2N focuses on residual learning, where the network learns only the noise rather than recovering the clean image, simplifying training and better preserving spectral information [35–37].

$$L_{res.}(\theta) = \frac{1}{2}\left(\left\|D_1(y) - f_\theta(D_1(y)) - D_2(y)\right\|_2^2 + \left\|D_2(y) - f_\theta(D_2(y)) - D_1(y)\right\|_2^2\right) \quad (3)$$

$$L_{cons.}(\theta) = \frac{1}{2}\left(\left\|D_1(y) - f_\theta(D_1(y)) - D_1(y - f_\theta(y))\right\|_2^2 + \left\|D_2(y) - f_\theta(D_2(y)) - D_2(y - f_\theta(y))\right\|_2^2\right) \quad (4)$$

The network ($f_\theta$) architecture is a simple two-layer image-to-image network, consisting of two convolutional operators with a 3×3 kernel size, followed by a 1×1 convolution. While deep learning models often require kernel size adjustments based on input image size, we avoid additional hyperparameter tuning by using this simple structure, maintaining stable performance. The small kernel size also allows future modifications to input image size. In summary, our proposed denoising method involves first processing noisy multi-spectral PA images through the SDDR module to generate a single noisy image $y$. This image is then fed into the SPADE framework for two-step denoising (BM3D and ZS-N2N). Finally, the denoised image $y'$ is fed into the reverse SDDR module to recover the denoised multi-spectral PA imaging.

### c. Implementation Detail and Evaluation Metric

In the proposed SPADE framework, all modules are integrated and packaged together for an end-to-end implementation. Data processing is conducted on a server equipped with two Intel Xeon Silver 4314 CPUs (2.40 GHz), two NVIDIA A100 80G GPUs, and 256 GB of RAM [38]. To evaluate the performance of our method, we introduced several quantitative metrics. One of the key measures is the signal-to-noise ratio (SNR) [39], which is used to assess the denoising capability. SNR is defined by Eq. 5:

$$SNR = 20 log_{10} \frac{|P_{peak}|}{\sigma_{background}} \quad (5)$$

where $P_{peak}$ is the peak PA signal amplitude and $\sigma_{background}$ is the standard deviation of background noise. The noise region was selected at the same depth as the peak signal, positioned laterally more than $1\ mm$ away from each peak.

We also used peak signal-to-noise ratio (PSNR) [35] and structural similarity (SSIM) index [21,40] to quantify the similarity between the denoised image and the reference image. PSNR is calculated based on the image ($I$) and reference images ($I_{ref}$), as defined in Eq. 6. Both images ($I$ and $I_{ref}$) have $m \times n$ dimension where $m, n$ is the pixel coordinates.

$$PSNR = 10 log_{10} \frac{(P_{peak})^2}{\sqrt{\frac{1}{MN}\sum_{m=0}^{M-1}\sum_{n=0}^{N-1}(I_{ref}-I)^2}} \quad (6)$$

SSIM is another metric to evaluate imaging quality highlighting structural similarity. SSIM is defined in Eq. 7. $\mu_{ref}$ and $\mu_I$ are the mean of the reference image and denoised image, respectively. And $\sigma_{ref}$ and $\sigma_I$ are the standard deviation of the reference image and denoised. $\sigma_{ref,I}$ is the covariance between two images, and $c_1$ and $c_2$ are constant values to avoid instability when the sum square of means or variances are close to zero.

$$SSIM = \frac{(2\mu_{ref}\mu_I+c_1)(2\sigma_{ref,I}+c_2)}{\sqrt{(\mu_{ref}^2+\mu_I^2+c_1)(\sigma_{ref}^2+\sigma_I^2+c_2)}} \quad (7)$$

Additionally, we quantitatively evaluated spectrum information preservation by calculating Pearson correlation coefficients between spectra to assess changes in pixel spectrum.

### 3. Experimental Setup

#### a. Simulation Implementation

We first conducted a simulation study to statistically understand the performance of our proposed method. This study has two main objectives. First, to analyze the image SNR improvement performance of the proposed method across different noise levels while preserving spectral information. The second is to assess the coefficient of the spectral similarity after denoising to validate that spectral information is preserved. The simulation runs on a Matlab (Matlab 2023a, MathWorks, USA) platform with K-Wave [41] and ValoMC [42] for the time-domain PA simulation [43]. A 2D simulation environment with a space of 41.6 ×41.6 $mm^2$ was defined. 128 transducer elements were placed apart from each other at a distance of 0.195 $mm$. We used the default omnidirectional setting for the sensor and the sensor is a broadband point detector. Light source propagation simulation (optical simulation) is based on the finite element method (FEM) and the Monte Carlo method. A single Gaussian light source aligned with the transducer has been set up with a standard deviation of 0.1. One point target (diameter: 650 $\mu m$) was placed at a depth of 3.9 $mm$ with respect to the transducer plane.

In order to obtain multiple frame images, we independently run the light source propagation simulation several times. For multi-spectrum image simulation, we run the Light source propagation simulation several times with different absorption coefficients to mimic different spectra. The simulated PA signal first underwent beamforming using Delay-and-Sum (DAS) algorithms. Then, the different levels of noise are added at each simulation to simulate acoustic noise. The simulated single-frame noisy PA image was processed with vanilla BM3D and our proposed methods. The performance matrices with different methods were compared.

### b. Phantom Study

We then conducted a wire phantom study to assess the performance of the proposed method and verify its consistency with the simulation study. A phantom study used Nylon fish wire as the imaging target, which was affixed onto a 3D-printed wire holder. The phantom comprised four layers of vertically aligned wires and three-column layouts, facilitating the assessment of PA imaging quality along both axial and lateral axes. Vertical wires were positioned at depths of 8, 13, 18, and 23 $mm$ from the transducer surface, while three horizontal wires were situated at depths of 13 $mm$ and 18 $mm$, all with a diameter of 0.2 $mm$.

For imaging, the Verasonics Vantage system (Vantage 128, Verasonics, USA) was employed for data acquisition. A laser system (Phocus MOBILE, OPOTEK, USA) capable of emitting wavelength-tunable laser light (690 to 950 $nm$) at a repetition rate of 20 Hz with a 5 $ns$ pulse duration served as the light source. A linear transducer (L12-5, Philips, Netherlands) facilitated both US and PA imaging. In the phantom study, 700 to 850 $nm$ wavelength range was utilized for PA excitation, with US images acquired at the same location to validate the target structure. The acquired PA signal was beamformed using DAS algorithm. The acquired noisy PA image was processed with vanilla BM3D, averaging filter, and our proposed methods. The performance matrices with different methods were compared.

### c. Ex vivo Evaluation with Swine Cardiac Tissue

We also conducted an ex vivo study to analyze the performance of the proposed method using ex vivo tissue samples. The proposed SPADE framework was evaluated in clinically relevant scenarios, specifically for denoising sPA images in cardiac ablation monitoring. Intraoperative sPA imaging for cardiac scans presents challenges due to the dynamic cardiac motion, which limits the use of averaging filters. SPADE's single-frame denoising method is well-suited to address these challenges. The same experiment devices in the phantom study were used in this ex vivo study. For the imaging target, swine cardiac tissue was selected as the experimental substrate, with a total of 35 samples prepared and subjected to scanning procedures. The tissue samples were ablated with various durations to create ablation-induced necrosis lesions. The spectral information of the tissue was analyzed for functional guidance to identify ablated tissue regions under various denoising methods. The tissue samples were immersed in a water bath for acoustic coupling, and both US and PA images were acquired at the same location using the same imaging setup. PA image was acquired with a wavelength range between 700 to 850 $nm$ covering the spectral signature of normal tissue and ablated necrosis [44]. The performance of the methods in denoising and spectrum preservation was evaluated. The image denoised with 64 frames averaging filter was used as a ground truth, and single-frame noisy images were processed with vanilla BM3D and our proposed SPADE method.

### d. In vivo Demonstration

We also conducted an in vivo study to verify the performance of the proposed method on in vivo animal model. The objective of this study is verifying the performance of the proposed method in an in vivo environment. Respiratory motion and myoclonus (involuntary muscle contractions) during anesthesia can reduce the effectiveness of averaging filters. Additionally, using fewer frames for denoising allows for an increased imaging frame rate, which is crucial for real-time intraoperative guidance. The in vivo investigation was conducted utilizing a mouse tumor model employing sublines of a human prostate cancer cell line, originally derived from an advanced androgen-independent bone metastasis. Male non-obese diabetic (NOD)/severe combined immunodeficient (SCID) mice aged six to eight weeks were subcutaneously implanted near the posterior flanks. A total number of 11 mice were utilized in this study. A gold nanoparticle (AuNR) contrast agent, designed to attach to the tumor and enhance tumor detection in PA imaging, was synthesized and administered to the animals. The targeted contrast exhibited a PA spectrum peak at 780 $nm$. This study was approved by the Institutional Animal Care and Use Committee (IACUC) at Worcester Polytechnic Institute (Protocol number: 21-127).

Ultrasound was utilized for tumor localization at baseline, utilizing the same imaging setup as in the phantom study. The mice were anesthetized with 3% isoflurane in oxygen via a nose cone and positioned dorsally on a fixation stage. Ultrasound gel was applied to ensure proper coupling between the probe and the tumors, as illustrated in Fig. 2. Ultrasound B-mode images aided in transducer positioning and optimizing the interrogated tumor cross-section in PA imaging. After establishing the ultrasound slice position, consecutive PA images within the same image field were acquired, with a wavelength of 780 $nm$ selected to highlight the contrast agent's peak spectrum and visualize the targeted tumor.

An identical imaging setup was used to scan the animals. The wavelength between 700 to 850 $nm$ was used with an intermediate step of 10 $nm$. The acquired image was processed with vanilla BM3D, averaging filter, and our proposed methods. In addition, the processed sPA images were spectrally unmixed to reveal the distribution of the contrast agent in tissue and around the tumor. Animals were imaged before, and 24 hours post injection to visualize the changes in contrast agent distribution.

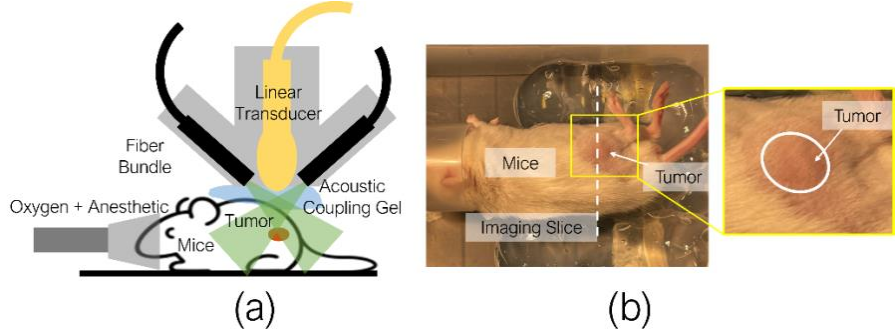

Figure 2. In vivo tumor scanning: (a) Animal setup sketch, (b) Photograph of prepared animal with tumor region zoomed in.

## 4. Results

### a. Simulation Implementation

The proposed SPADE denoising algorithm was evaluated through simulations using a grid of point targets with varying levels of random noise. Its performance was compared to the vanilla BM3D method, with results shown in Fig. 3(a). The raw input image reveals significant noise around the point targets. After processing with BM3D, noise is reduced, but residuals remain, particularly in deeper regions. In contrast, the SPADE-processed image reveals more uniform noise suppression, with clearer, more consistent point targets at all depths.

Quantitative evaluations used four metrics: SNR, spectrum similarity measured by correlation coefficient, SSIM, and PSNR. SPADE consistently achieved higher denoised SNR across a wide range of input SNRs as shown in Fig. 3(b), excelling when the input SNR was below 25 $dB$. BM3D also improved SNR but was less effective in low-SNR conditions, only matching SPADE at higher input SNRs. The Δ SNR results suggest that SPADE reaches superior performance in high-noise environments, with an average SNR improvement exceeding 15 $dB$ for input SNRs below 25 $dB$, far surpassing BM3D. Both methods reached a performance plateau as input SNR increased, with fewer differences at higher SNRs.

Spectrum similarity, measured by the correlation coefficient between the denoised and ground truth spectra, was also evaluated in Fig. 3(c). SPADE maintained a higher correlation across noise levels, especially at lower input SNRs, indicating better spectral preservation. Both methods performed well (R > 0.8) at higher SNRs, but SPADE demonstrated greater stability, with higher average coefficients and lower variability compared to BM3D. Finally, the statistical analysis of SSIM and PSNR in Fig. 3(d) further emphasizes SPADE's advantages. The box plots indicate that SPADE achieved higher scores on both metrics with less variability, demonstrating its consistent performance across different noise levels.

In addition to comparing the denoising performance of the proposed SPADE method, we evaluated its effectiveness across different input wavelength numbers to validate its generality. PA images with 2 to 15 wavelengths were derived from the original 16-wavelength simulated sPA data and input into the SPADE method. The same evaluation metrics, SNR improvement, SSIM, and PSNR, were used to assess effectiveness at various input dimensions (see Fig. 3(e)). The quantitative data demonstrate stability across the number of input wavelengths. Average SNR improvement increases from 13.32 $dB$ at 2 wavelengths to 14.19 $dB$ at 16 wavelengths. SSIM improves from 0.838 to 0.846, and PSNR rises from 42.21 $dB$ to 44.49 $dB$.

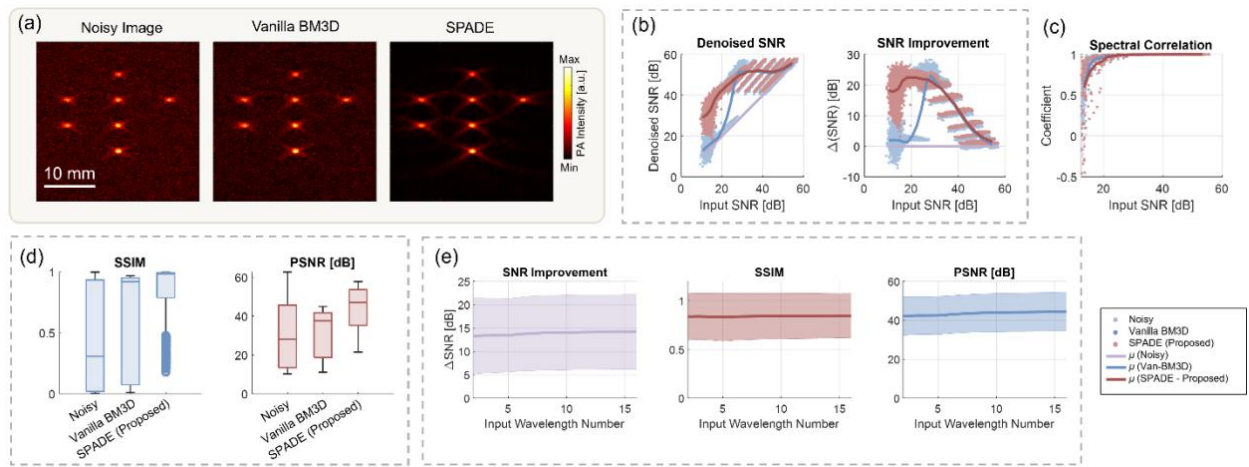

Figure 3. Simulated spectroscopic photoacoustic (sPA) image at 720 $mm$. (a) PA images (left to right): noisy single-frame PA image, vanilla BM3D denoised image, and image denoised by the proposed SPADE method. Quantitative evaluation of the SPADE method compared to vanilla BM3D and the noisy single-frame PA image: (b) Output signal-to-noise ratio (SNR) at different noise levels and SNR improvement. (c) Spectral correlation coefficient of vanilla BM3D and SPADE at various noise levels. (d) Structural similarity (SSIM) index and Peak signal-to-noise ratio (PSNR), computed with respect to the ground truth simulation image without noise. (e) Denoising performance across varying numbers of input wavelengths.

### b. Phantom Study

A phantom study using a multi-point target, similar to the simulation, was conducted to evaluate the real-world effectiveness of the proposed SPADE denoising method. The denoised images are presented in Fig. 4, where the noisy image was processed using a 64-frame averaging filter, vanilla BM3D, and the proposed SPADE method.

Fig. 4(a) illustrates the visual results of the denoising process. The noisy image exhibits reduced visibility of the point targets, particularly at lower depths. After applying the 64-frame averaging filter, some noise suppression is achieved, but the reduction is uneven across depths. The vanilla BM3D method further improves noise reduction, though some residual noise remains, especially at the deepest points. In contrast, the SPADE method suppresses the noise more uniformly, with high contrast point targets of preserved shape across the entire image.

Fig. 4(b) and (c) quantitatively evaluate the denoising performance. The box plot shows the distribution of denoised SNR values for each method, with SPADE consistently yielding higher median SNR values and a tighter interquartile range, indicating more reliable performance across varying noise levels. The central plot examines denoised SNR as a function of input SNR, where SPADE demonstrates substantial improvement, particularly at lower input SNRs where other methods struggle. SPADE's ability to maintain a higher SNR across all noise levels underscores its effectiveness in enhancing image quality under challenging conditions. Fig. 4(d) assesses spectral similarity, with all methods showing a strong correlation (R > 0.99).

Fig. 4(e) and (f) further support these findings through additional statistical analysis. The box plots compare SSIM across methods, where the proposed SPADE method reaches a lower mean SSIM compared to the averaging filter and vanilla BM3D. Its standard deviation is larger than BM3D's but smaller than the averaging filter's. A higher PSNR is observed with the proposed method compared to vanilla BM3D, with a comparable standard deviation, indicating consistent performance relative to the averaging filter.

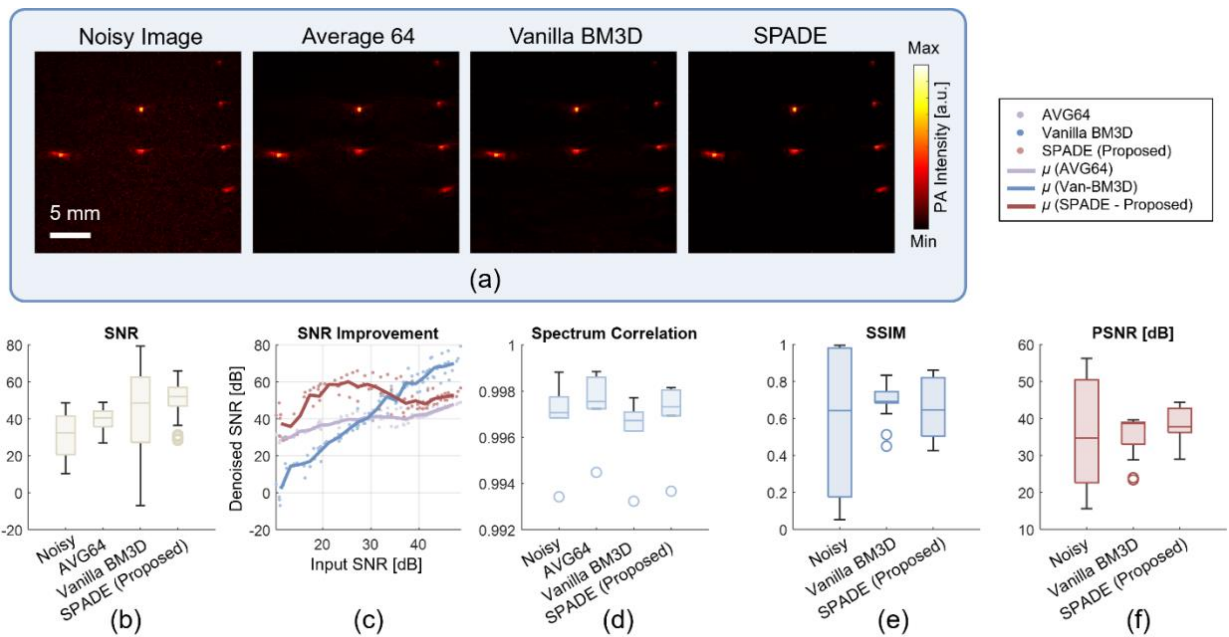

Figure 4. (a) Photoacoustic (PA) imaging denoising results of the point phantom at 810 $nm$. From left to right: Raw noisy data, 64-frame averaging filter, vanilla BM3D, and the proposed SPADE algorithm. (b) SNR statistics of point targets for each method. (c) Output SNR from the three denoising methods at various noise levels. (d) Spectral correlation coefficient of each method compared to the ground truth spectrum. (e) Structural similarity (SSIM) index and (f) Peak signal-to-noise ratio (PSNR) computed relative to the averaging filter image.

### c. Ex vivo Evaluation with Swine Cardiac Tissue

The proposed methods were then evaluated using ex vivo swine cardiac tissue. Sixteen wavelengths of PA scanning were performed on 35 ablated cardiac tissue samples. The single-frame noisy image was processed using vanilla BM3D and the proposed SPADE method, with the 64-frame averaging filter result used as the ground truth for comparison. Fig. 5(a) shows the denoised image results from each method. In the single-frame noisy image, the tissue boundary is barely visible, with its location confirmed via ultrasound. All three denoising methods successfully suppressed noise, highlighting both the tissue boundary and the ablated region. The SPADE method revealed more intensity from deeper regions compared to vanilla BM3D. Spectral unmixing was performed on the PA images using spectra from ablated and non-ablated tissue [44], displaying the ablation-induced necrotic boundary. The noisy data failed to identify tissue boundaries due to electrical noise above the tissue surface, and the boundary of the ablated tissue was inaccurate. In contrast, all three denoising methods provided a clear boundary, matching the gross pathology results acquired post-scanning. The tissue and lesion shapes obtained by the SPADE method align better with vanilla BM3D.

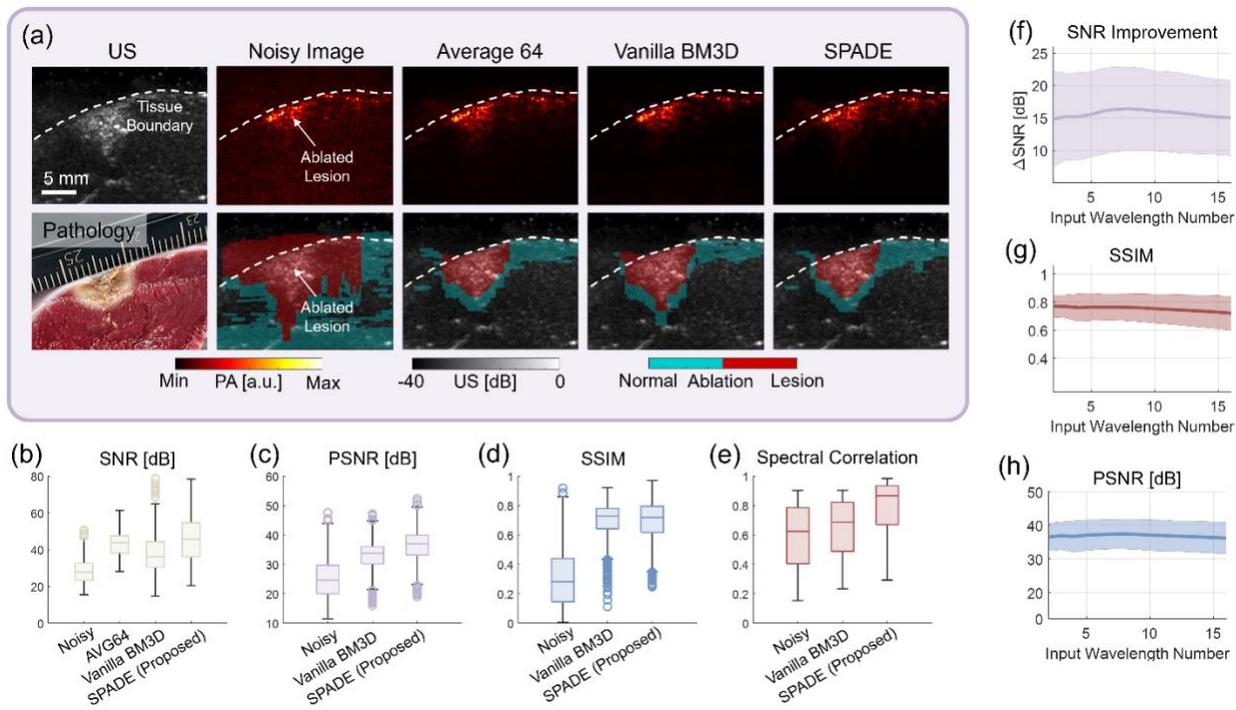

Figure 5. Ex vivo evaluation of spectroscopic PA (sPA) analysis for cardiac ablation mapping. (a) PA image at 780 $nm$ and PA-mapped lesion boundary using different denoising methods, alongside the corresponding US image and post-scan gross-pathology highlighting the lesion. Quantitative analysis statistics: (b) Signal-to-noise ratio (SNR), (c) Peak signal-to-noise ratio (PSNR), (d) Structural similarity (SSIM) index, and (e) spectral correlation coefficient relative to the averaging filter. Denoising performance: SNR improvement (f), SSIM (g), PSNR (h) across varying numbers of input wavelengths based on ex vivo evaluation.

Statistical analysis of all ex vivo scans was presented. SNR values (Fig. 5 (b)) directly reflect the denoising effectiveness of each method. The proposed SPADE method achieved the output (43.05 $dB$ median) from the original 28.01 $dB$ noisy image, close to the 64-frame averaging filter (43.22 $dB$ median). Both PSNR (Fig. 5(c)) and SSIM (Fig. 5(d)) values suggest that all denoising methods improved image structural similarity, with the SPADE method yielding a higher PSNR (36.173 $dB$) compared to vanilla BM3D (32.632 $dB$) and a comparable SSIM (SPADE: 0.697, vanilla BM3D: 0.721). The spectral correlation from the SPADE denoised image was higher, indicating strong agreement with the ground truth spectra from the averaging filter image, as shown in Fig. 5(e).

We evaluated the effectiveness of SPADE denoising across different input wavelengths based on ex vivo results. Wavelengths ranging from 2 to 15 were derived from the original images and input into the SPADE method. In Fig. 5(f-h), the same evaluation metrics were used to assess effectiveness at various input dimensions. The results indicate that average SNR improvement varies between 14.80 $dB$ and 16.42 $dB$ across different input wavelengths. SSIM remains relatively stable with a slight decline from 0.771 to 0.721, while PSNR remains steady between 36.17 $dB$ and 37.37 $dB$.

d. **In vivo Demonstration**

In vivo demonstration was conducted using a contrast agent-injected mouse model to enhance tumor detection. The same imaging and spectral analysis pipeline from the previous ex vivo study was applied. The spectrum of the AuNR agent acquired from the earlier study was used as the library spectra, and the scanned sPA images were unmixed with AuNR, oxygenated hemoglobin (HbO2), and deoxygenated hemoglobin (HbR) spectra. The denoised PA images and contrast agent distribution maps, both before and 24 hours after injection, are shown in Fig. 6. The contrast agent distribution was quantified by calculating the ratio of its decomposed pixel intensity to the total hemoglobin intensity (HbR + HbO2), accounting for local illumination variation.

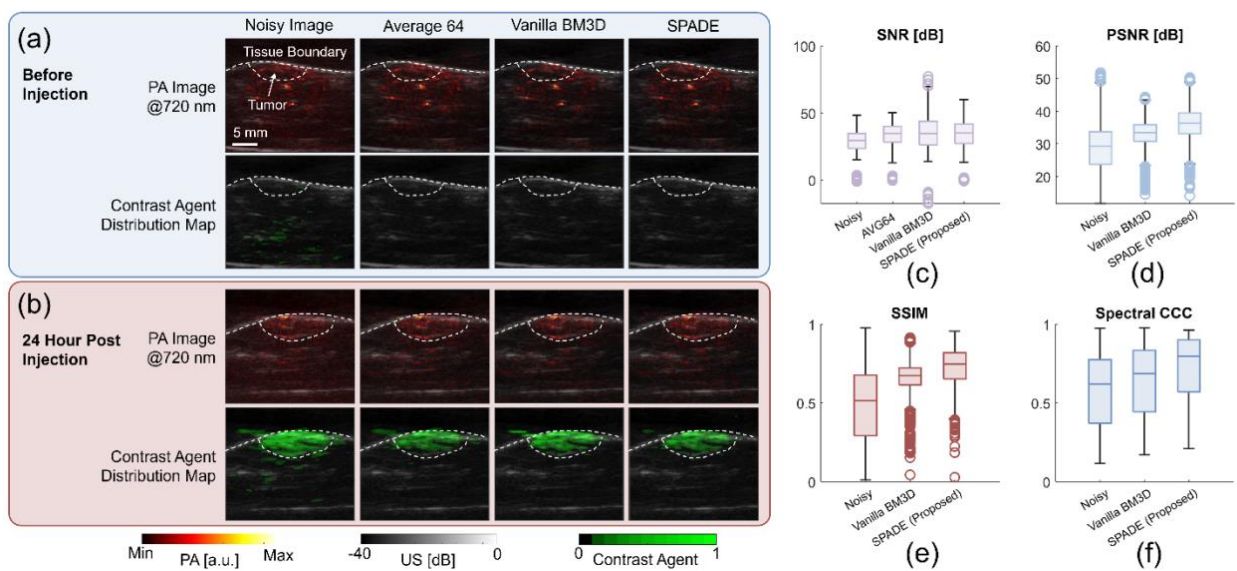

Figure 6. In vivo mice tumor scanning: (a) before and (b) 24 hours after contrast agent injection. From left to right: single frames, PA image averaged with 64 frames (Reference), vanilla-BM3D, and SPADE (proposed), overlaid with the US image. The same color range is applied within each method before and after injection. Quantitative analysis statistics: (c) Signal-to-noise ratio (SNR), (d) Peak signal-to-noise ratio (PSNR), (e) Structural similarity (SSIM) index, and (f) spectral correlation coefficient relative to the averaging filter.

As in the ex vivo results, all three denoising methods effectively suppressed noise and enhanced tissue boundaries that were otherwise invisible in the noisy image. Before injection, no contrast agent signals were detected across all denoising cases, as depicted in Fig. 6(a). In the noisy image, some false contrast agent intensity values were observed in the deeper regions due to mislabeled noisy spectra. 24 hours after injection, a strong contrast agent signal was observed around the tumor region as presented in Fig. 6(b). In the noisy image, the contrast agent was mainly distributed in the tumor region, with some illuminated pixels in the deeper regions. In the denoised results, the contrast agent intensity value was predominantly retained within the tumor region. The distribution of AuNR in the proposed method closely matched the averaging filter results. Notably, the SPADE method was the only result without noisy pixels illuminated outside the tissue boundary.

Statistical analysis is presented in Fig. 6(c-f). The median SNR of the SPADE method is 35.09 $dB$, comparable to the averaging filter at 34.86 $dB$, and higher than both vanilla BM3D at 34.71 $dB$ and the original noisy image at 29.50 $dB$. The PSNR between the SPADE method and the averaging filter is 36.26 $dB$, an improvement over the noisy image (29.28 $dB$) and vanilla BM3D (33.47 $dB$). A similar trend is observed in SSIM, with SPADE achieving 0.748 compared to 0.514 from the noisy image and 0.673 from BM3D. The spectral correlation of the denoised image with the ground truth spectrum from the averaging filter was also evaluated. The SPADE method achieved a median coefficient of 0.798, higher than 0.620 before denoising, and 0.686 with vanilla BM3D.

## 5. Discussion

The results from the simulation, phantom, in vivo, and ex vivo studies consistently demonstrate the superior performance of the proposed SPADE denoising algorithm in quantitative PA imaging compared to vanilla BM3D and achieving comparable performance to 64 frames averaging filters which generally considered as ground truth. Across all experiments, SPADE was able to achieve more uniform noise suppression, improved SNR, and enhanced image quality, particularly in the challenging low-SNR environments. These improvements are particularly significant given the limitations of existing denoising techniques, which often struggle in high-noise conditions and result in less reliable performance in deeper regions.

In the simulation study, SPADE consistently outperformed BM3D in terms of noise suppression and quantitative metrics such as SNR, SSIM, and PSNR. Notably, SPADE can better preserve the clarity of the point target across different depths, particularly in low-SNR conditions, whereas BM3D's performance is weakened. This trend was reflected in the statistical evaluations, where SPADE maintained a higher average SNR and correlation coefficient, especially when the input SNR fell below 25 $dB$. This finding underscores the robustness of SPADE in handling noise without sacrificing the fidelity of the spectroscopic information, as evidenced by its performance in spectrum similarity preservation. The evaluation of denoising performance across various input wavelengths highlights the stability and robustness of the SPADE framework. The results suggest that SPADE maintains effective denoising even with as few as two input wavelengths, underscoring its versatility and efficiency in handling different input dimensions without compromising performance.

The phantom study further corroborates these findings, demonstrating that SPADE achieves higher SNR values and more effective noise suppression in a physical model. Compared to the 64-frame averaging filter and BM3D, SPADE provided clearer point targets across varying depths. This result indicates the method's usability in scenarios where high frame averaging may be impractical or introduce latency. Additionally, SPADE's performance consistency across different noise levels, as illustrated by the tighter interquartile range in SNR values, reinforces its applicability in diverse imaging conditions. Although the median SSIM value for SPADE (0.647) is lower than that of vanilla BM3D (0.691), and a similar trend is seen with PSNR, SPADE suppressed more noise around the point targets, as confirmed by both SNR measurements and visual inspection. The lower SSIM and PSNR values are likely due to the removal of signal sidelobes in SPADE images, which are present in the averaging filter images due to beamforming.

The ex vivo evaluation using swine cardiac tissue verifies the effectiveness of SPADE in clinical settings. Here, the algorithm demonstrated a strong capacity to enhance tissue boundary visibility, which was challenging the noisy data. The quantitative spectral analysis, which accurately revealed the ablation-induced necrotic boundary, further confirms the spectral preservation of our proposed method. Statistical analysis showed that SPADE achieved the highest median SNR and PSNR values, closely aligning with the ground truth from the 64 frames averaging filter. The ability of SPADE to maintain higher spectral correlation coefficients with single-frame noisy data input indicates its potential for increasing PA imaging speed which is critical for future clinical translation. The performance evaluation across different input wavelengths aligns with the simulation results, validating the robustness of the SPADE framework.

In vivo results on contrast agent-enhanced tumor detection provided additional confirmation of SPADE's applicability in living animals. The algorithm successfully suppressed noise while retaining the integrity of the contrast agent quantification, which was observed in both before and 24 hours after injection scanning. Images denoised by SPADE closely matched the reference 64-frame averaged images and outperformed BM3D in terms of all the quantitative matrices. Importantly, SPADE was the only method that completely eliminated false contrast agent signals outside the tissue boundary, demonstrating its precision in preserving relevant features while removing artifacts. This capability is particularly valuable in vivo imaging applications where accurate quantification of molecular and structural targets is critical for diagnosis and therapy monitoring.

Although the results successfully demonstrate the effectiveness of the SPADE method in denoising sPA images, several limitations exist. First, the denoising was performed using PA images collected from the same hardware, and the generalizability of the method across different imaging setups and transducers has not been tested. Therefore, evaluation using various imaging probes is necessary. Secondly, the current denoising process using SPADE is time-consuming, even with high computational power. Denoising a 16-wavelength sPA image takes approximately 70 seconds, comparable to the duration of an 87.5-frame average using a 20 Hz repetition laser. While real-time visualization is not yet feasible, the method eliminates the need for motion-gating during averaging in cases of dynamic cardiac or respiratory motion. Future efforts to optimize the algorithm and network to improve processing speed are needed. Thirdly, the introduction of a learning-based N2N network relies on stochastic gradient descent, introducing randomness to the output even with identical input [45,46]. This randomness leads to higher standard deviation in performance metrics. While no instability in spectral outcomes was observed in this study, a more robust evaluation is needed to assess potential performance variation due to this randomness. The effectiveness of the SPADE framework in suppressing noise arising from electromagnetic interference, such as MRI [47] or radiofrequency ablation energy [48], was not evaluated in this study and requires future investigation.

## 6. Conclusion

In conclusion, the proposed SPADE framework effectively addresses the challenges of denoising sPA imaging. By combining a data-free learning-based method with an analytical approach, SPADE achieves higher noise suppression while preserving spectral information, critical for quantitative PA imaging. Through extensive validation in simulation, phantom, ex vivo, and in vivo experiments, the method exhibited significant improvements in SNR compared to noisy images, with comparable outcomes to the averaging filter. The framework consistently provided clearer image visualization and more accurate spectral quantification, confirming its potential for enhancing sPA imaging in various clinical applications. Future work should focus on optimizing processing speed and assess the generality of the method across different imaging systems for real-time clinical implementation.

**Acknowledgements**

This work was supported by Worcester Polytechnic Institute Internal Fund and the National Institutes of Health under grants: R01DK133717, DP5OD028162, R01CA134675 R01CA166379 and R01EB030539. We thank the Worcester Polytechnic Institute Animal Care and Use Committee for their assistance with this study.